\documentclass[sigconf]{acmart}
\AtBeginDocument{%
  }

\setcopyright{acmlicensed}
\copyrightyear{2024}
\acmYear{2024}
\acmDOI{XXXXXXX.XXXXXXX}

\acmConference[CONSEQUENCES]{Consequences 2024}{October 2024}{Bari, Italy}

\acmJournal{JDS}
\acmVolume{37}
\acmNumber{4}
\acmArticle{111}
\acmMonth{8}

\usepackage{algpseudocode}
\usepackage{algorithm}
\usepackage{amsmath}

\begin{document}

\title{Batched Online Contextual Sparse Bandits with Sequential Inclusion of Features}

% \email{andrew@metica.com}

\author{Rowan Swiers, Subash Prabanantham, Andrew Maher}
\affiliation{%
    \country{London UK,} \institution{Metica}
}

% \author{Rowan Swiers, Subash Prabanantham, Andrew Maher}
% % \affiliation{%
% %   \country{London, UK}
% %   \institution{Metica, London, UK} 
% % }
% \affiliation{%
%   \institution{Metica} 
%   \country{London, UK}
% }

\email{{rowan, subash, andrew}@metica.com}

\renewcommand{\shortauthors}{Swiers et al.}
%%
%% Article type: Research, Review, Discussion, Invited or position
\acmArticleType{Review}
%%
%% Links to code and data
% \acmCodeLink{https://github.com/borisveytsman/acmart}
% \acmDataLink{htps://zenodo.org/link}
%%
%% Authors' contribution
% \acmContributions{BT and GKMT designed the study; LT, VB, and AP
%   conducted the experiments, BR, HC, CP and JS analyzed the results,
%   JPK developed analytical predictions, all authors participated in
%   writing the manuscript.}
%%
%% Sometimes the addresses are too long to fit on the page.  In this
%% case uncomment the lines below and fill them accodingly.
%%
%% \authorsaddresses{Corresponding author: Ben Trovato,
%% \href{mailto:trovato@corporation.com}{trovato@corporation.com};
%% Institute for Clarity in Documentation, P.O. Box 1212, Dublin,
%% Ohio, USA, 43017-6221}

% keywords can be removed

\begin{abstract}

Multi-armed Bandits (MABs) are increasingly employed in online platforms and e-commerce to optimize decision-making for personalized user experiences. In this work, we focus on the Contextual Bandit problem with linear rewards, under conditions of sparsity and batched data. We address the challenge of fairness by excluding irrelevant features from decision-making processes using a novel algorithm, Online Batched Sequential Inclusion (OBSI), which sequentially includes features as confidence in their impact on the reward increases. Our experiments on synthetic data show the superior performance of OBSI compared to other algorithms in terms of regret, relevance of features used, and compute.

\end{abstract}

\maketitle

\section{Introduction and Motivation}

The core functionality of MABs \cite{Lattimore2020BanditA,Slivkins2019IntroductionTM} is their ability to balance exploration and exploitation, by dynamically allocating traffic, to maximize cumulative rewards based on observed outcomes. Contextual MABs extend this by incorporating user-specific contextual information, enabling more refined and targeted decision-making.

Sparsity arises when only a subset of features within a context are relevant to the reward, while the remaining features are irrelevant to the reward. Addressing sparsity is not only crucial for optimizing total reward, but it also has significant implications for the fairness of decisions made with a MAB. For instance, in a hiring decision process, incorporating irrelevant features, such as a person's clothing, could be unfair. 

In practice, MABs are frequently deployed in batch settings where multiple actions are taken before rewards are received and the model can be updated, e.g. in clinical trials\cite{Robbins1952SomeAO,Cheng2005BayesianAD} and marketing campaigns.

This paper focuses on online sparse batched contextual MABs, aiming to improve both efficiency and fairness by ensuring that only relevant features are used in decision-making. We propose the Online Batched Sequential Inclusion OBSI algorithm, which effectively handles sparse contexts while maintaining strong performance across fairness and regret metrics.

\section{Related Work} % \label{Results}

For low dimensional bandits Agrawal et al. \cite{Agrawal2012ThompsonSF} introduced Thompson Sampling that maintains a posterior distribution of the rewards. The OLS bandit was introduced in \cite{Goldenshluger2013ALR}  which introduced forced sampling to find the right balance of exploration and exploitation.

In the domain of Sparse bandits one approach is the "Optimism in the Face of Uncertainty" \cite{AbbasiYadkori2012OnlinetoConfidenceSetCA}, which utilizes a confidence set-based method for handling sparse bandits. This approach is similar to ours in that it can be applied to different algorithms.  Several other approaches have been proposed, Bastani et al \cite{Bastani2015OnlineDW} employed the Lasso estimator \cite{Tibshirani1996RegressionSA} combined with forced sampling, to achieve a low regret bound. Wang et al. \cite{Wang2018MinimaxCP} proposed the MCP Bandit, which leverages the MCP estimator \cite{Zhang2010NearlyUV}, a less biased alternative to Lasso , and again used forced sampling. Kim et al. \cite{Kim2019DoublyRobustLB} introduced the Doubly Robust Bandit, which uses the Lasso Doubly Robust estimator \cite{Bang2005DoublyRE}, avoiding the need for forced sampling and requiring fewer tuning parameters.

In the context of greedy bandit algorithms, Bastani \cite{Bastani2017MostlyEA} discussed the effectiveness of greedy strategies, particularly in low-dimensional settings.

Regarding batched learning, Han et al. \cite{Han2020SequentialBL} examined both adaptive and fixed-size batches, showing that pure exploitation can be highly effective in low-dimensional contexts.  Kalkanli et al \cite{Kalkanli2021AsymptoticPO} showed the optimality of batched Thompson sampling.  Similarly, Ren et al. \cite{Ren2020DynamicBL} utilized the Lasso estimator and demonstrated that a pure exploitation approach performs well in sparse batched settings.

Fairness in machine learning has garnered increasing attention in recent years \cite{Verma2018FairnessDE,Dwork2011FairnessTA}. Numerous fair bandit algorithms have been proposed \cite{Joseph2016FairnessIL,Liu2017CalibratedFI}, each adopting different definitions and approaches to fairness.

 \section{Method} %\label{Related work}

\subsection{Overview}

Our agent is designed to solve the contextual MAB problem with linear rewards under sparsity and batched data. We also introduce a fairness regret, which is incurred when actions are influenced by irrelevant features. The agent learns the confidence in the impact of each individual feature and only incorporates features into the decision-making process once a confidence threshold is met.

\subsection{Formulation}

\sloppy An agent $\pi$ interacts with an environment over \( n \) rounds. In round \( t \in [n] \), the agent observes a context \( X_t \in \mathbb{R}^d \), chooses an action \( A_t \in \mathcal{A} \) and receives a reward $R_t \in \mathbb{R}$. The data is received in batches defined by a grid $\Gamma = \{t_i\}^M_{i=0}$ with $t_0=0$ and $t_M=n$. The rounds in batch $k$ are $\{t|t_{k-1}<t \leq t_k\}$. The rewards for a particular action and context are not observed by the agent until the end of a batch. The batched history is a replay of previous events before the current batch $H_t =(X_1,A_1,R_1,...,X_{j(t)},A_{j(t)},R_{j(t)})$ where $j(t)=\max( \{t_i | t_i \in \Gamma , t_i \leq t\})$. The chosen actions depends on the batched history and the context. $A_t \sim \pi(A_t|X_t,H_{t-1})$. The reward has distribution  \( R_t \)=$X_t^\top\theta_A + s \epsilon_t$, where the $\epsilon_t$ are IID Gaussian noise and $s \in \mathbb{R}$ is a parameter.

The context \( X \) can be split into two parts: \( X_S \) and \( X_N \), where \( X_S \) represents relevant features (signal) that influence the reward, and \( X_N \) represents irrelevant features (noise) that do not affect the reward. The agent does not know beforehand which features are relevant or irrelevant. So for some $i$ the latent parameters $\theta_A^i=0$ for all $A$. 

A standard performance measure of a bandit algorithm is the regret.  If we let  \( \mu(A, X) = X^\top\theta_A \) and $A^*_t = \arg \max_{A \in \mathcal{A}} \mu(A, X_t)$, then the $n$-round regret is defined as:

\begin{equation}
    \text{Regret}_n(\pi) = \mathbb{E}\left(\sum_{t=1}^{n} \mu(A_t^*, X_t) - \mu(A_t, X_t)\right)
    \label{eq:regret}
\end{equation}

We also measure the fairness of our bandit algorithm by looking at how much the irrelevant features affect the action choice. The fairness  regret measures the probability that an action choice changes due to irrelevant features and is measured at each batch using MC sampling. If we let $X_S \sim N(0,I)$ and $X_N^{(i)} \sim N(0,I)$ to make $X^{(i)}=(X_S,X_N^{(i)})$. Then $A^{(i)}_t \sim \pi(A^{(i)}_t |X_t^{(i)},H_{t-1})$ and the fairness regret is defined as:

\begin{equation}
    Fairness_M(\pi) = \sum_{i=1}^{M} P(A^{(1)}_{t_i}  \neq A^{(2)}_{t_i})
    \label{eq:fairness}
\end{equation}

\subsection{Algorithm}

We suppose the reward $R_t$ is generated according to $R_t \sim N(X _t^\top\theta_{A}, s^2I).$ Given a prior that the reward is distributed as  $N(\boldsymbol{0}, v^2I)$. We define the following quantities:

\begin{equation*}
B_{t,A} = I_d + \sum^{t-1}_{\tau=0}X^\top_\tau X_\tau1(A_\tau=A)
\end{equation*}

\begin{equation*}
\hat{\theta}_{t,A}=B_{t,A}^{-1}(\sum^{t-1}_{\tau=0} X_\tau^\top R_\tau)1(A_\tau=A)
\end{equation*}

After receiving a reward $R_t$, and assuming a prior $P(\theta_t) \sim N(\theta_{t}, v^2 B_{t}^{-1})$, the posterior $P(\theta_{t+1} |R_t) \propto P(R_t| \theta_t)P(\theta_t)$ is distributed as $N(\theta_{t+1}, v^2 B_{t+1}^{-1})$. These are used in Algorithm \ref{alg:obsi} to efficiently estimate the posterior, in an online fashion, by updating $B_t$ and $\hat{\theta_{t}}$ at each time step.

One problem with batched linear Thompson Sampling BLTS \cite{Kalkanli2021AsymptoticPO} applied to sparse data is that features are used to make decisions before they are known to have an impact on the reward. This increases the fairness regret as irrelevant features are used to make a decision. It also increases regret due to over fitting. Hence, we define $\Bar{\theta}^i_{t,A}$ in equation (\ref{eq:sequential-inclusion}), which sets $\tilde{\theta}^i_{t,A}$ to zero in the sampling stage if there is low confidence that the true value of $\theta_i$ is non-zero. This leads to Algorithm \ref{alg:obsi} which is the description of the online batched sequential inclusion algorithm OBSI.

\begin{equation}
\Bar{\theta}^i_{t,A} =
\begin{cases} 
      \tilde{\theta}^i_{t,A} & \text{if } \dfrac{\sum_{A \in \mathcal{A}  } |\hat{\theta}_{t,A}^i|}{\sqrt{\sum_{A \in \mathcal{A}} \text{Var}(\hat{\theta}_{t,A}^i)}}>\Phi^{-1}(\alpha)\\
      0 & \text{otherwise }  
\end{cases}
\label{eq:sequential-inclusion}
\end{equation}

\begin{algorithm}
\caption{Online Batched Sequential Inclusion (OBSI)}
\begin{algorithmic}
    \State \textbf{Input:} Batch size $M$, grid $\Gamma = \{t_i\}^M_{i=0}$
    \For{$i = 1,...,M$}
        \For{$t = t_i, ..., t_{i+1}$}
            \State Sample $\tilde{\theta}_{t,A} \sim N(\hat{\theta}_{t,A}, v^2 B_{t_i,A}^{-1})$ 
            \State Calculate $\Bar{\theta}$ using $\tilde{\theta}$ and (\ref{eq:sequential-inclusion})
            \State Set $A_t = \arg\max_{A \in \mathcal{A}} X_t^\top \bar{\theta}_{t,A} $
        \EndFor
        
    \EndFor
\end{algorithmic}
\label{alg:obsi}
\end{algorithm}

 A key benefit of the algorithm is ability to share information between $\theta_A$ coefficients. Additionally, the tuneable $\alpha$ parameter allows for precise control over how quickly contexts are incorporated as their relevance becomes more certain. Finally, the concept of sequential inclusion of features can be applied to other bandit algorithms, provided a posterior distribution is available.

\section{Experiments} \label{Results}

A synthetic dataset was created with the following settings: $M = 40$, $n=20{,}000$; $|\mathcal{A}|=5 $; $d = 20$, $|X_S| =10$, $|X_N|=10$, $X \sim N(0,I)$. Each batch consisted of 500 rounds. In each run $\theta_{t,A}^i \sim N(0,1)$ and for the non relevant features $\theta_{t,A}^i=0$ for all $A$. The reward function is $R_t \sim N(X _t^\top\theta_{A}, 10^2I)$.

We compare our algorithm, OBSI, against: (1) the \textit{Oracle} method, Batched Linear Thompson Sampling from alg. \ref{alg:BLTS}, trained only on the relevant features $X_c$, (2) BLTS trained on all features in $X$; (3) Lasso Batch Greedy Learning (LBGL) \cite{Ren2020DynamicBL} (4) MCP batched greedy \cite{Wang2018MinimaxCP}. Pseudo code for the comparison algorithms are available in the appendix. Compute time was measured on a 2023 Apple M2 Pro, with 32GB of ram and the simulations were repeated 1000 times.  

The results, summarized in Table \ref{tab:comparison}, indicate that OBSI outperforms both the LBGL and MCPB approaches in terms of computation time, owing to its fully online nature which eliminates the need for retraining complex estimators like Lasso and MCP. Furthermore, while excluding the Oracle method, OBSI achieves a lower regret and a lower total fairness regret compared to other baseline methods. These results highlight the efficiency and effectiveness of OBSI in environments with moderate levels of sparsity.

\begin{table}[]
\centering
\resizebox{\columnwidth}{!}{%
\begin{tabular}{llll}
\hline
\textbf{Algorithm}   & \textbf{Regret (Eq. \ref{eq:regret})} &  \textbf{Fairness (Eq. \ref{eq:fairness})} & \textbf{Compute {[}s{]}}  \\ \hline
Oracle      & 7.8       & 0.00         & 18.3         \\ \hline
BLTS        & 12.7      & 8.11       & 19.3         \\ \hline
LGBL        & 10.8      & 5.70         & 103         \\ \hline
MCPB         & 12.0      & 7.94         & 168         \\ \hline
OBSI (ours) & \textbf{9.8}       & \textbf{2.70}         & \textbf{18.4}         \\
\end{tabular}%
}
\caption{Comparison of MABs in terms of regret, fairness and compute. For all three metrics, lower is better.}
\label{tab:comparison}
\vspace{-\baselineskip}
\vspace{-\baselineskip}

\end{table}

\section{Conclusion}

In conclusion, our approach, OBSI, demonstrated significant improvements over existing methods, particularly in scenarios characterized by a moderate level of sparsity. OBSI is notably efficient, capable of being fully executed in an online manner without the need for iterative retraining. Our experiments on synthetic datasets confirmed that OBSI effectively reduces regret and also reduces fairness regret compared to other methods.

Future work could further explore the application of the sequential inclusion approach, potentially applying it to more complex real-world datasets. 

%\nocite{*}
\bibliographystyle{ACM-Reference-Format}
\bibliography{samplebib}

\appendix

\section{Appendix}

\subsection{Comparison of Regret Evolution}

\begin{figure}
  \centering
  \includegraphics[width=\linewidth]{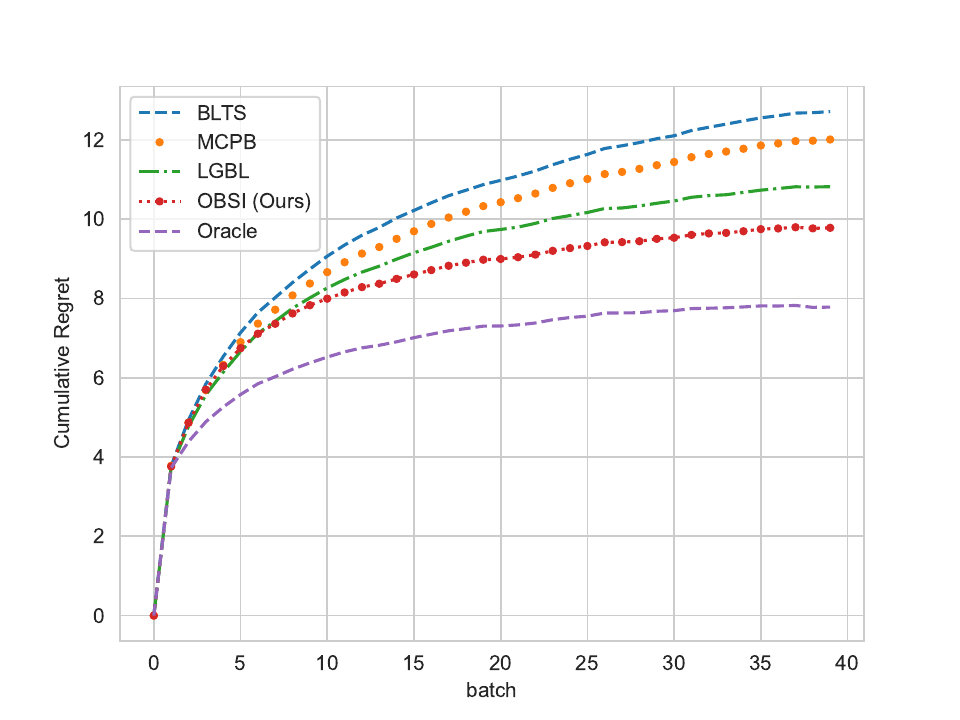}
  \caption{Regret of bandit algorithms }
  \label{fig:regret_exp}
\end{figure}

Figure \ref{fig:regret_exp} illustrates the evolution of regret for the different bandit algorithm as the number of batches increases. The rate of increase of cumulative regret slowly decreases as more batches are included. OBSI outperforms the other methods at each time step.

\subsection{Comparison of Alpha Values}

\begin{figure}
  \centering
  \includegraphics[width=\linewidth]{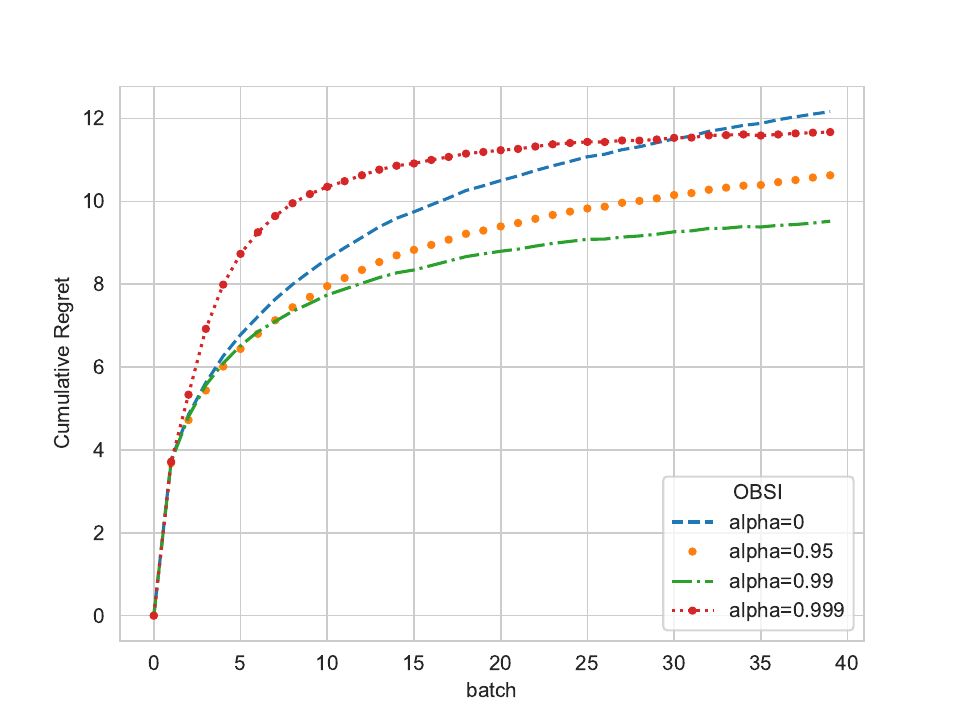}
  \caption{Regret for different alpha values }
  \label{fig:alpha_regret}
\end{figure}

\begin{figure}
  \centering
  \includegraphics[width=\linewidth]{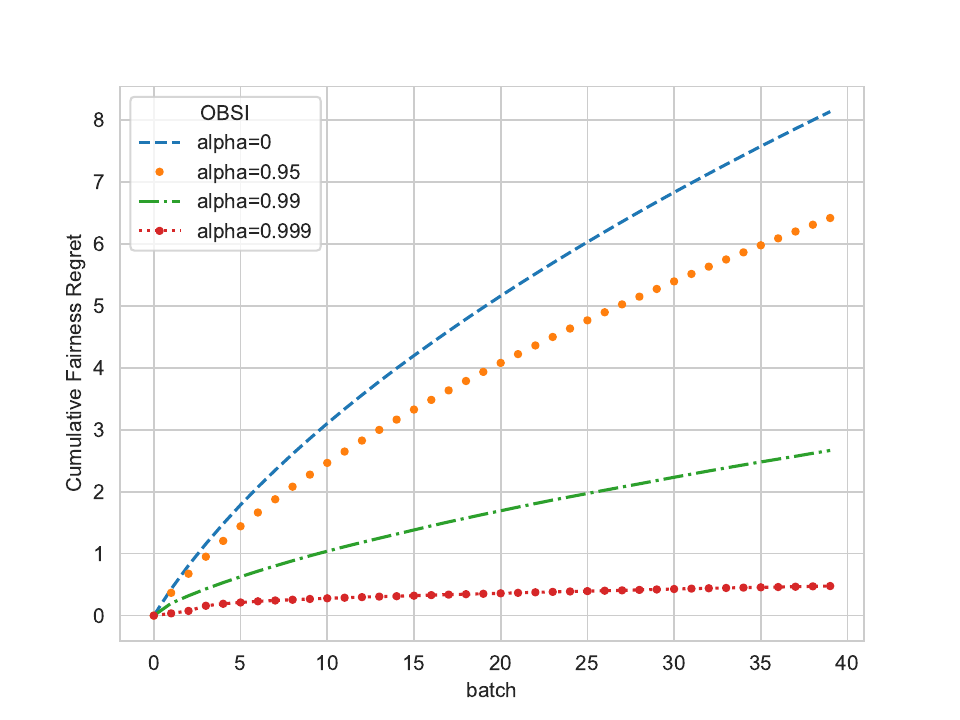}
  \caption{Fairness Regret for different alpha values }
  \label{fig:alpha_fair}
\end{figure}

We evaluated the performance of the OBSI algorithm across various values of the hyper-parameter $\alpha$. The regret associated with these values is depicted in Figure \ref{fig:alpha_regret}. The results indicate that there is an optimal range for setting $\alpha$. When $\alpha$ is too strict, the algorithm under-utilizes features, while a more lenient $\alpha$ leads to over-fitting. Figure \ref{fig:alpha_fair} illustrates the fairness regret across different $\alpha$ thresholds. As the $\alpha$ threshold increases, the fairness score consistently improves, showing that by tuning $\alpha$ the fairness can be controlled. 

\subsection{Comparison of Different Dimensions}

\begin{figure}[H]
  \centering
  \includegraphics[width=\linewidth]{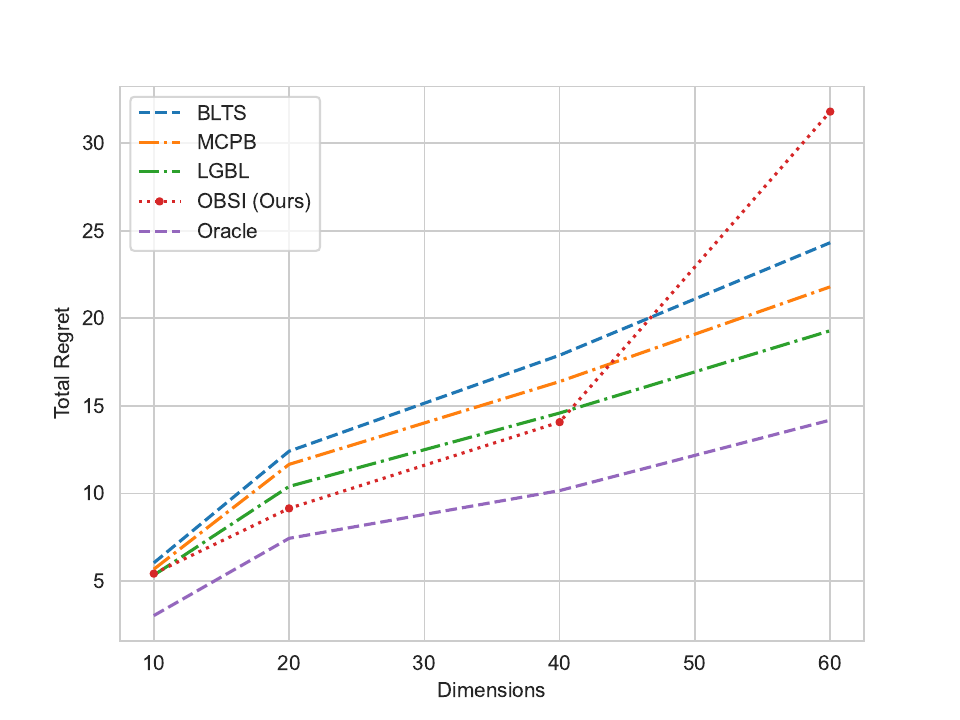}
  \caption{Regret for different dimensions }
  \label{fig:dimension_comparison}
\end{figure}

We evaluated the effect of varying the number of dimensions on the algorithm's regret.  We kept an equal ratio of relevant to non-relevant contexts as we changed the number of dimensions. Figure \ref{fig:dimension_comparison} shows the results of the experiment. The OBSI method demonstrated superior performance compared to other methods when the number of dimensions was fewer than 40. However, as the number of dimensions exceeded 40, the performance of OBSI declined relative to the other methods. This may be because an appropriate confidence level can not be reached to include a context feature.

\subsection{Comparison Algorithm PseudoCode}

We present the pseudo code for the comparison algorithms used in our study. Algorithm \ref{alg:BLTS} outlines the Batched Linear Thompson Sampling (BLTS) algorithm\cite{Kalkanli2021AsymptoticPO,Agrawal2012ThompsonSF}. This algorithm is particularly suited for low-dimensional settings. Algorithm \ref{alg:LGBL} presents the LASSO Batched Greedy Learning (LGBL) \cite{Ren2020DynamicBL,Bastani2015OnlineDW} algorithm, while Algorithm \ref{alg:MCP} describes the MCP bandit (MCPB) \cite{Zhang2010NearlyUV} algorithm. Hyper-parameters were tuned via grid search. Given the relatively large batch sizes in our experiments and due to the measurement of fairness, forced sampling was not implemented for these approaches.

\begin{algorithm}[H]
\caption{Batched Linear Thompson Sampling}
\begin{algorithmic}
    \State \textbf{Input:} Number of batches $M$, grid $\Gamma = \{t_i\}^M_{i=0}$
    \For{$i = 1,...,M$}
        \For{$t = t_i, ..., t_{i+1}$}
            \State Sample $\tilde{\theta}_{t,A} \sim N(\hat{\theta}_{t,A}, v^2 B_{t_i,A}^{-1})$ 
            \State Choose $A_t = \arg\max_{A \in \mathcal{A}} X_t^\top \tilde{\theta}_{t,A} $
        \EndFor
    \EndFor
\end{algorithmic}
\label{alg:BLTS}
\end{algorithm}

\begin{algorithm}[H]
\caption{LASSO Batch Greedy Learning (LBGL)}
\begin{algorithmic}
    \State \textbf{Input:} Number of batches $M$, grid $\Gamma = \{t_i\}^M_{i=0}$
    \For{$i = 1,...,M$}
        \For{t = $t_i, ..., t_{i+1}$}
            \State Choose $A_t = \arg \max_{A \in A} X_{t}^\top \hat{\theta}_{m-1,A} $ 
        \EndFor
        \State $\hat{\theta}_{m,A} = \arg \min_{\theta} \frac{1}{t_{m}} \sum_{t=0}^{t=t_m} (R_{t,A_t} - X_{t,A_t}^ \top \theta)^2 + \lambda_m || \theta||_1   $ 
    \EndFor
\end{algorithmic}
\label{alg:LGBL}

\end{algorithm}

\begin{algorithm}[H]
\caption{MCP Batch Greedy (MCPB)}
\begin{algorithmic}
    \State \textbf{Input:} Number of batches $M$, grid $\Gamma = \{t_i\}^M_{i=0}$
    \For{$i = 1,...,M$}
        \For{t = $t_i, ..., t_{i+1}$}
            \State Choose $A_t = \arg \max_{A \in A} X_{t}^\top \hat{\theta}_{m-1,A} $ 
        \EndFor
        \State $\hat{\theta}_{m,A} = \arg \min_{\theta} \frac{1}{t_{m}} \sum_{t=0}^{t=t_m} (R_{t,A_t} - X_{t,A_t}^ \top \theta)^2 + \sum_{j=1}^dP_{\lambda,a}(\theta^j)   $ 
    \EndFor
\end{algorithmic}
\label{alg:MCP}

\end{algorithm}

Where $P_{\lambda,a}(\theta^j) = \int_{0}^{|\theta^j|} \max(0,\lambda-\frac{1}{a}|t|) \mathrm{d} t$

\end{document}